\documentclass[runningheads]{llncs}
\usepackage{booktabs}
\usepackage{graphicx}
\usepackage{orcidlink}
\usepackage{tabularx}
\usepackage{multirow,xcolor}
\newcolumntype{Y}{>{\centering\arraybackslash}X}
\newcolumntype{L}{>{\hsize=.85\hsize\centering\arraybackslash}X} 
\newcolumntype{R}{>{\hsize=1.2\hsize\centering\arraybackslash}X} 

\newcolumntype{S}{>{\hsize=.65\hsize\centering\arraybackslash}X} 
\newcolumntype{D}{>{\hsize=1.8\hsize\centering\arraybackslash}X} 

\usepackage{enumitem}
\usepackage{bm}
\newcommand{\specialcell}[2][c]{%
\begin{tabular}[#1]{@{}c@{}}#2\end{tabular}}
\DeclareUnicodeCharacter{2212}{\ensuremath{-}}
\newcommand{\Lagr}{\mathcal{L}}
\usepackage{float}
\usepackage{amsmath}
\usepackage[misc]{ifsym}
\begin{document}
\title{Gaussian Pancakes: Geometrically-Regularized 3D Gaussian Splatting for Realistic Endoscopic Reconstruction}
%
%
\author{Sierra Bonilla\inst{1,2}
 \and
Shuai Zhang\inst{1,3} \and
Dimitrios Psychogyios\inst{1,2} \and
Danail Stoyanov\inst{1,2} \and
Francisco Vasconcelos\inst{1,2} \and
Sophia Bano\inst{1,2}}

%
\authorrunning{S. Bonilla et al.}
\titlerunning{Gaussian Pancakes}

\institute{Wellcome/EPSRC Centre for Interventional and Surgical Sciences (WEISS), UK \email{\{sierra.bonilla.21,shuai.z\}@ucl.ac.uk} \and
Department of Computer Science, University College London, London, UK \and
Department of Medical Physics and Biomedical Engineering, University College London, UK}

\maketitle             
\begin{abstract}
Within colorectal cancer diagnostics, conventional colonoscopy techniques face critical limitations, including a limited field of view and a lack of depth information, which can impede the detection of precancerous lesions. Current methods struggle to provide comprehensive and accurate 3D reconstructions of the colonic surface which can help minimize the missing regions and reinspection for pre-cancerous polyps. Addressing this, we introduce ``Gaussian Pancakes'', a method that leverages 3D Gaussian Splatting (3D GS) combined with a Recurrent Neural Network-based Simultaneous Localization and Mapping (RNNSLAM) system. By introducing geometric and depth regularization into the 3D GS framework, our approach ensures more accurate alignment of Gaussians with the colon surface, resulting in smoother 3D reconstructions with novel viewing of detailed textures and structures. Evaluations across three diverse datasets show that Gaussian Pancakes enhances novel view synthesis quality, surpassing current leading methods with a 18\% boost in PSNR and a 16\% improvement in SSIM. It also delivers over 100$\times$  faster rendering and more than 10$\times$ shorter training times, making it a practical tool for real-time  applications. Hence, this holds promise for achieving clinical translation for better detection and diagnosis of colorectal cancer. 
Code: \url{https://github.com/smbonilla/GaussianPancakes}.

\end{abstract}
\section{Introduction}

Colorectal cancer remains a major global health challenge, consistently ranking among the top three cancers in prevalence and mortality \cite{WorldHealthOrganization_2022}. The slow progression of colorectal cancer offers detecting pre-cancerous growth through colonoscopy. Despite several AI-based methods proposed in the literature for polyp detection during colonoscopy \cite{barua2020artificial}, challenges persist, particularly in identifying polyps hidden behind folds, which account for up to three-quarters of all missed cases~\cite{pickhardt2004location}. Colonoscopists encounter difficulties in thorough surface inspection, due to the use of a monocular camera with limited field of view and lack of 3D information while navigating the colon's complex structure. Generating full 3D reconstructions with high-quality textures from endoscopic images in near real-time would enable improved diagnosis and treatment through downstream tasks like automated surgical path planning \cite{9712211}, AR/VR training environments\cite{sun2023dynamic}, AI-based polyp detection \cite{barua2020artificial}, and for determining missing regions for re-inspection \cite{ma2021rnnslam}.  

The exploration of novel view synthesis and 3D reconstruction in endoscopy has significantly advanced with the advent of Neural Radiance Fields (NeRF)~\cite{mildenhall2020nerf} and Neural Implicit Surfaces (NeuS)~\cite{wang2023neus} which learn a continuous function that implicitly represents the 3D scene trained from 2D images and paired camera poses. However, NeRF and NeuS,  including their colonoscopy-specific adaptations ColonNeRF\cite{shi2023colonnerf}, REIM-NeRF\cite{psychogyios2023realistic} and LightNeuS\cite{batlle2023lightneus}, are limited by lengthy training times, slow rendering speeds, and a lack of explicit geometry, which complicates integration into practical workflows. Both NeRF and NeuS rely on pre-known camera poses and depths, further restricting their versatility. To address these challenges, the field of computer vision and surgical vision\cite{huang2024endo,liu2024endogaussian,zhuendogs} has recently shifted its attention to 3D Gaussian Splatting (3D GS)\cite{kerbl20233d}. This technique represents 3D scenes through populations of 3D Gaussians, offering real-time rendering, explicit manipulable geometry for easier integration, and significantly reduced training times. Nonetheless, it faces challenges, including a lack of geometry constraints leading to artifacts in untrained views and the necessity of a robust point cloud and known camera poses as starting points.

\begin{figure}[t!]
\centering
\includegraphics[width=0.9\textwidth]{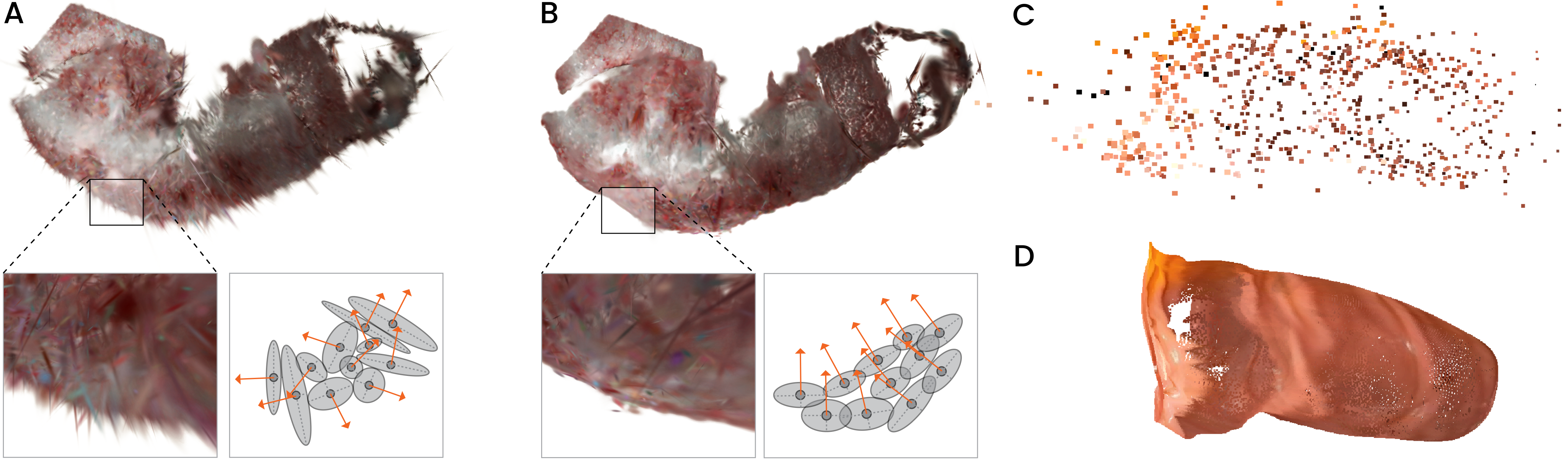}
\caption{Illustrating the benefit of Gaussian Pancakes, arrows indicating Gaussians' principal normal direction, A) Gaussian Splatting without Pancaking (PSNR = 37.54) and B) Gaussian Splatting with Pancaking (PSNR = 38.50), on a synthetic video sequence generated by a colonoscopy simulator~\cite{zhang2020template}; C) Sparse point cloud from SfM and D) point cloud from RNNSLAM on the In-Vivo dataset~\cite{ma2021rnnslam}.}\label{normals}
\end{figure}

In response to these limitations, we propose Gaussian Pancakes, an approach that synergizes a geometrically-regularized 3D GS with a colonoscopy-tailored Simultaneous Localization and Mapping (SLAM) system, RNNSLAM \cite{ma2021rnnslam}, to enhance texture rendering and anatomical accuracy while boosting training and rendering speeds. Our method capitalizes on the strengths of RNNSLAM, which generates camera poses, depth maps, and rudimentary surface reconstruction from only the monocular video inputs in real-time, providing a robust alternative to traditional Structure-from-Motion (SfM) methods. Although RNNSLAM alone yields surface reconstructions, the lack of photorealism and anatomical detail limits its use in novel view synthesis tasks. Gaussian Pancakes overcomes these drawbacks by replacing surfels with 3D Gaussians and incorporating surface-based regularization, resulting in significantly accurate radiance fields and vivid textures. Our key contributions are: \textbf{1)}~A pipeline that integrates 3D GS with SLAM for photorealistic 3D colonoscopic reconstructions, demonstrating robustness over using a SfM method seen in Fig.~\ref{normals} (C-D); \textbf{2)}~Improved 3D GS method by incorporating geometric and depth regularizations, aligning Gaussians with the surfaces by effectively ``pancaking'' them. This enhancement reduces geometric error and artifacts in novel view synthesis, seen in Fig.~\ref{normals}; \textbf{3)}~Optimized training and rendering process for radiance fields in surgical scenes, reducing training times to roughly 2 minutes and improving image rendering speeds by more than 100 times, making it more applicable in clinical settings.

\section{Method}

\begin{figure}[t!]
\includegraphics[width=0.9\textwidth]{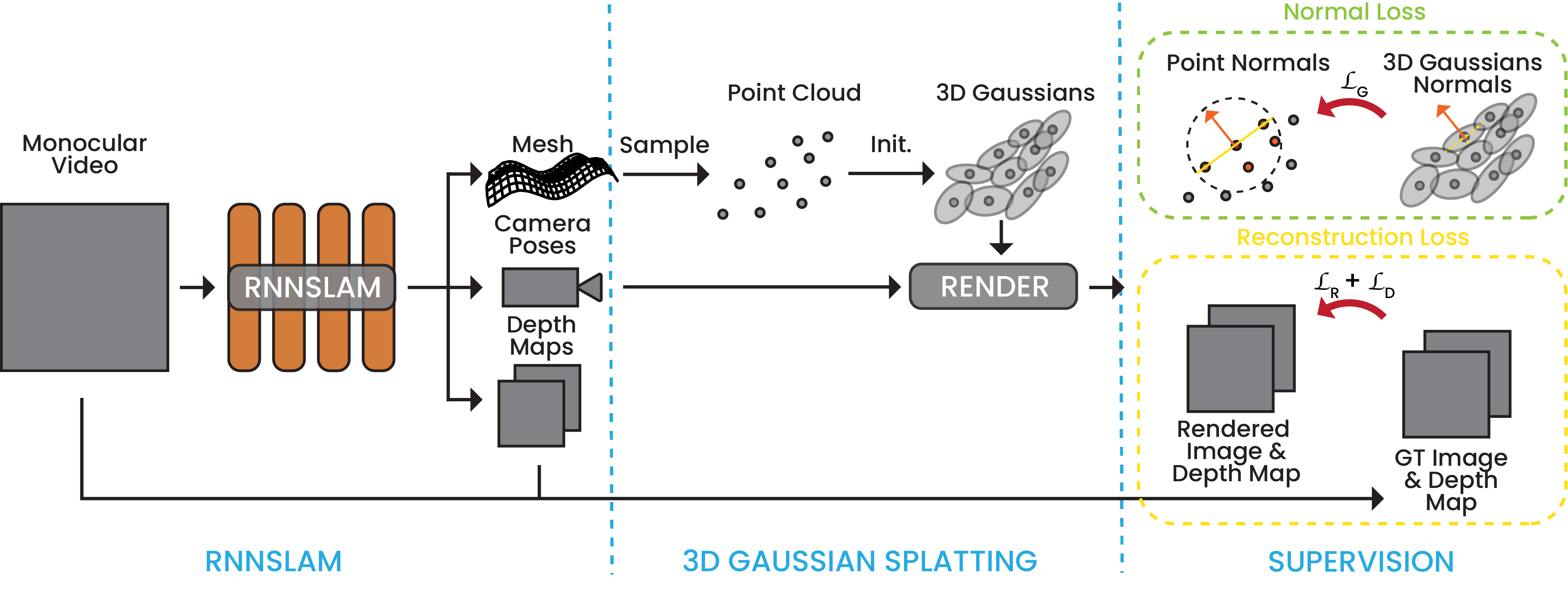}
\centering
\caption{Proposed Gaussian Pancakes' pipeline highlighting our contributions: A) RNNSLAM for mesh, camera poses, \& depth maps; B) 3D GS initialization, C) Geometric \& depth regularizations, distinguishing our approach from traditional 3D GS.} \label{method}
\end{figure}

Gaussian Pancakes employs 3D GS for realistic texture rendering
and RNNSLAM for 3D reconstruction. As shown in Fig .~\ref{method}, it starts with RNNSLAM-generated camera poses, depth maps, and meshes. This mesh is sampled, using stratified sampling, to create a sparse RGB
point cloud \cite{alexa2004stratified}. Around these points, Gaussians with corresponding attributes are initialized (Sec. 2.2). Supervision is twofold: photometric and geometric regularization, with the addition of geometric regularization ensuring Gaussians align with the surface and limit floating artifacts.

\subsection{Preliminaries of RNNSLAM}
RNNSLAM \cite{ma2021rnnslam} uses sparse depth estimation from a SfM method as a proxy for ground truth to train a recurrent neural network for depth and camera pose prediction. Bundle-adjusted direct sparse odometry is used to jointly optimize the predicted poses and depth by minimizing the intensity difference over a window of recent frames. The fusion pipeline is then used to reconstruct colon meshes with textures, enabling the detection of missing regions. While this approach has shown effectiveness in reconstructing dense 3D colon maps from visible sections in real colonoscopy, RNNSLAM suffers from pose estimation drift, resulting in texture misalignment and structural collapse in the fused meshes, which is a common problem among similar methods \cite{rau2023simcol3d}. Our proposed method uses the refined depth maps, camera poses and sparsely sampled points on converged meshes of RNNSLAM to provide high-quality texture renderings of the structures. 

\subsection{Preliminaries 3D Gaussian Splatting}
3D GS employs Gaussian distributions for scene representation and leverages GPU-accelerated rendering for efficient optimization \cite{kerbl20233d}, \cite{lassner2021pulsar}, \cite{chen2024survey}. 3D GS method models scenes using 3D Gaussians around the input RGB point cloud, represented by position $\bm{\mu}$, scale $s$, orientation as quaternion $\mathbf{q}$, spherical coefficients $SH$, and opacity $\sigma$. Each Gaussian has a covariance matrix $\Sigma$ describing the skew and variance of the distribution in 3 dimensions, calculated using the scaling matrix $S$ and rotation matrix $R$ by 
\begin{equation}
\Sigma = RSS^{T}R^{T}
\label{sigma_eq}
\end{equation}
A Gaussian can be described by the multivariate Gaussian probability density:
\begin{equation}
G(\mathbf{x}) = e^{−\frac{1}{2}(\mathbf{x}−\bm{\mu})^{T}\Sigma^{-1}(\mathbf{x}−\bm{\mu}))}
\end{equation}
These Gaussians are expressed in 3D space but are projected onto 2D for image rendering given by a viewing transformation $W$ and the covariance matrix $\Sigma$. The covariance matrix in the camera coordinate system is given by $\Sigma ' = JW\Sigma W^{T}J^{T}$, where $J$ is the Jacobian matrix of the affine approximation of $W$. The splatting process involves differentiable point-based alpha blending. The final image pixel color is accumulated from contributions along a ray, considering the densities and transmittances. The color of each Gaussian is represented using spherical harmonic coefficients, capturing view-dependent effects \cite{ramamoorthi2001efficient}. The Gaussian attributes are optimized using the Adam optimizer \cite{kingma2017adam} through a rendering process, combined with steps to adjust Gaussian density.

\subsection{Proposed Gaussian Pancakes: Extending 3D GS for Endoscopy}

Instead of relying on the sparse point cloud output from an SfM method, our approach utilizes RNNSLAM for pose estimation, depth maps, and RGB point cloud generation \cite{schoenberger2016sfm}. 
This substitution addresses SfM's limitations in low-variation environments where uniform textures can impede accurate feature matching and depth perception, thoroughly described in the RNNSLAM paper \cite{ma2021rnnslam}. 

\noindent \textbf{Depth regularization:} Endoscopic images have minimal diversity in image directions, causing 3D GS to overfit. This results in artifacts where depth is misrepresented, representing images onto a 2D plane \cite{chung2024depthregularized}. To mitigate this problem, we introduce a depth loss based on the Huber loss. For the depth maps $\{D_{i}\}_{i=1}^{M}$, with $M$ is the number of viewpoints, the depth loss $\Lagr_{D}(i)$ is defined as:
\begin{equation}
\Lagr_{D}(i) =
\left\{
    \begin{array}{lr}
        0.5\Delta  D_{i}^{2}, & \text{   if } |\Delta D_{i}| < \delta\\
        \delta (\Delta D_{i} - 0.5\delta), & \text{otherwise } 
    \end{array}
\right\}
\label{deptheq}
\end{equation}
$\Delta D_{i} = |D_{i}-\hat{D}_{i}|$ denotes the loss at $i$-th frame, and $\hat{D}_{i}$, the rendered depth. 

\noindent \textbf{Geometric regularization:} Despite the depth loss, challenges persist due to the lack of surface constraints. Artifacts which present as floating objects often emerge as the rendered view gets further away from original camera positions. To mitigate this, we incorporate a geometric loss that constrains the Gaussians in accordance with the surface's principal curvature, effectively ``pancaking'' the Gaussians along the surface. For a point cloud \begin{math} P = \{\mathbf{p}_{m}\} \end{math} where \begin{math} \{\mathbf{p}_{m}\} = (x_{m}, y_{m}, z_{m}) \end{math} represents the $m$-th point, we compute the normal vector $\mathbf{n}_p$ for each point $\mathbf{p}_{m}$ in $P$. This computation assumes the local neighborhood of any point can be approximated by a tangent plane inferred from its 10 nearest neighbors. We denote these neighbors as \begin{math} \{\mathbf{n}_{j}\}_{j=1}^k \end{math}, where $k$ = 10. Centering the nearest neighbors by \begin{math} \mathbf{n'}_{j} = \mathbf{n}_{j} - \frac{1}{k} \sum_{j=1}^{k} \mathbf{n}_{j} \end{math}. The covariance of $\mathbf{n'}$, $C$, is calculated
\begin{equation}
C = \frac{1}{k-1} \sum_{j=1}^{k} \mathbf{n'}_{j} \mathbf{n'}^{T}_{j}
\end{equation}
Eigen decomposition of $C$ yields eigenvalues and eigenvectors $\mathbf{v}_{i}$. The normal vector, $\mathbf{n}_{p}$, for each point, $\mathbf{p}_{m}$, is the eigenvector corresponding to the smallest eigenvalue. The concatenated matrix of normal vectors and their points, $N$, is indexed using a GPU-accelerated structure, enhancing search efficiency \cite{johnson2017billionscale}.

Normal vectors for each 3D Gaussian are efficiently recalculated per iteration with a custom CUDA kernel for the dynamic scene. For each Gaussian's rotation matrix $R$, we find the axis of least spread by locating the minimum index of its scale vector $s$. This axis, selected from the standard Cartesian set ${e_x, e_y, e_z}$ as $\mathbf{e}_{min}$, is the primary basis for computing the normal vector $\mathbf{n}_{GP} = R \mathbf{e}_{min}$.
Using normals $A$ from the nearest point cloud points and Gaussian normals $B$, we calculate geometric loss with cosine similarity as follows: 
\begin{equation}
\Lagr_{G} = 1 - \Bigg | \frac{A\cdot B}{||A||\text{ }||B||}\Bigg | 
\end{equation}

\noindent \textbf{Loss Function:} Our model's total loss $\Lagr$ merges image reconstruction, depth, and geometric losses, balanced by weights $\lambda_{1}, \lambda_{2}, \lambda_{3}$. This compact formulation:
\begin{equation}
\Lagr = (1-\lambda_{1})\Lagr_{Image} + \lambda_{1}\Lagr_{D-SSIM} + \lambda_2 \Lagr_{D} + \lambda_3 \Lagr_{G}
\label{totalloss}
\end{equation}with $\Lagr_{Image}$ an L1 loss and $\Lagr_{D-SSIM}$, a structural similarity term, between rendered and ground truth images. This loss optimizes trade-offs between visual fidelity, depth error, and geometric alignment with empirically set weights.
\begin{table}[t!]
\caption{Assessment of photometric, reconstruction errors, training and rendering times against top methods across Simulation, Phantom, In-vivo datasets.}\label{tab1}

\begin{tabularx}{\textwidth}{R L L L R S S S}
\toprule
Model & Dataset & PSNR ↑ & \specialcell{SSIM ↑\\MS-SSIM*\verb| |} & \verb| |LPIPS ↓ & \specialcell{Depth ↓\\MSE} & FPS↑ &  \specialcell{GPU ↓\\min}\\
\toprule
ColonNerf\cite{shi2023colonnerf} &  Phantom & 25.54 &  0.86* &  0.40 & - & - & -\\
\multirow{3}{*}{NeRF\cite{mildenhall2020nerf}} & Simulation & 35.29 &  0.92 &  0.14 & 0.007 & $<$2 & 167 \\
& Phantom & 32.10 &  0.81 &   0.39  & 4.263 & $<$2 & 88 \\
& In-Vivo & 18.93 &  0.67 &  0.43 & 0.109 & $<$2 & 89\\
\multirow{3}{*}{ \specialcell{REIM-\\NeRF}\cite{psychogyios2023realistic}} & Simulation & 32.22 &  0.82 &  0.33 & $\bm{0.001}$ & $<$2 & 170 \\
& Phantom  &  31.66 &  0.78 &  0.22 & $\bm{0.013}$ & $<$2 & 120\\
& In-Vivo & 18.94 &  0.65 &  0.45 & $\bm{0.006}$ & $<$2 & 97\\
\multirow{3}{*}{\specialcell{Gaussian\\Pancakes\\(ours)} } & Simulation & $\bm{40.34}$ &  $\bm{0.97}$ &  $\bm{0.05}$  & 0.007 & $>$ $\bm{100}$ & $\bm{0.83}$ \\
 & Phantom & $\bm{32.31}$ &  $\bm{0.90/1.00*}$  &  $\bm{0.20}$ & 0.498 & $>$ $\bm{100}$ & $\bm{1.70}$ \\
& In-Vivo  & $\bm{26.25}$ &  $\bm{0.83}$ & $\bm{0.21}$ & 0.156 & $>$ $\bm{100}$ & $\bm{1.25}$\\
\toprule

\end{tabularx}
\end{table}

\begin{table}[!hbtp]
\caption{Ablation study on In-Vivo data.}\label{tab3}
\begin{tabularx}{\textwidth}{D S S S S}
\toprule
Method & PSNR ↑ & \specialcell{Depth ↑\\SSIM} & \specialcell{Depth ↓\\MSE} &  \specialcell{GPU ↓\\min}\\
\toprule
GS & 26.116 &  0.444 & 0.158 & 1.211  \\
GS+Depth& 26.062 & 0.446 & $\bm{0.145}$ & $\bm{1.206}$\\
GS+Pancaking & 26.211 & $\bm{0.460}$ & 0.154 & 1.329 \\
GS+Pancaking+Depth(ours) & $\bm{26.248}$ & 0.458 & 0.156 & 1.249\\
\toprule
\end{tabularx}
\end{table}

\section{Experiments}

\subsubsection{Dataset and Evaluation Metrics:}
Our evaluation encompasses three distinct datasets: Simulation \cite{zhang2020template}, In-Vivo \cite{ma2021rnnslam}, and Phantom \cite{bobrow2023colonoscopy}. The Simulation dataset, featuring sequences of the cecum, rectum, and sigmoid regions at a resolution of 320 $\times$ 240, is derived from software simulations in Unity using the graphics engine\cite{zhang2020template} for image rendering. Depth and pose references, however, are obtained from RNNSLAM. The In-Vivo dataset comprises real colonoscopy sequences each at 270 $\times$ 216 resolution from the RNNSLAM study\cite{ma2021rnnslam}. We selected three distinct sequences (\verb|"cecum_t4_b"|, \verb|"desc_t4_a"|, and \verb|"transt_t1_a"|) covering the cecum, descending colon and transcending sections from the public Phantom dataset (C3VD\cite{bobrow2023colonoscopy}), with 1350 $\times$ 1080 resolution colonoscopy videos. C3VD’s simple camera trajectories caused large gaps in reconstructions due to insufficient scene coverage for RNNSLAM’s mesh surface fusion step, limiting us to choose three datasets with satisfactory RNNSLAM performance for evaluation. 

Frame data for each scene is divided into training and testing sets with an 8:1 ratio, following the approach recommended in prior research \cite{barron2022mipnerf}. For a comprehensive quality assessment of comparative view synthesis, we use several common metrics: Peak Signal-To-Noise Ratio (PSNR), Learned Perceptual Image Patch Similarity (LPIPS), specifically the VGG loss \cite{zhang2018perceptual}, and Structural Similarity Index Measure (SSIM)/Multi-Scale Structural Similarity Index Measure (MS-SSIM) - for direct comparison with ColonNeRF. We measure Depth Mean Squared Error (MSE), frames per second (FPS) to evaluate rendering speed, and GPU minutes for training time (aggregated minutes across all GPUs used).

\noindent\textbf{Implementation:} For a fair comparison with REIM-NeRF and NeRF we downsampled the resolution of the phantom dataset (C3VD) to 338 $\times$ 270, similar to what was tested in \cite{psychogyios2023realistic}. We keep the original resolution of simulation and in-vivo datasets, since they are already close to \cite{psychogyios2023realistic}. Our methodology initiates with RNNSLAM, rendering 10 FPS on an Intel i7 CPU and Nvidia GeForce GTX 1080 GPU. Gaussian Pancakes uses inferred depth maps for supervision, along with inferred camera poses and points for initial setup, across all training sequences. Subsequently, Gaussian Pancakes models trained following \cite{kerbl20233d}'s hyperparameters, but are trained for 7,000 iterations instead of 30,000, densifying Gaussians until the 4,000th iteration. We set $\delta$ at 0.2 in Eq.~\ref{deptheq}, with hyperparameters in Eq.~\ref{totalloss} $\lambda_{1}, \lambda_{2}, \text{and } \lambda_{3}$ finely tuned to 0.2, 0.6, and 0.2 (empirically set), respectively, balancing photometric, reconstruction, and geometric results. Geometric loss integration commences after the first 1000 iterations for non-uniform scaling. All models are trained for 1-2 minutes on an NVIDIA RTX A6000.

\section{Results and Discussion} In Table \ref{tab1}, Gaussian Pancakes demonstrates performance improvements in photometric errors PSNR, SSIM, and LPIPS, and achieves over 100 FPS rendering speeds and training times less than 2\% that of the other models for all datasets. Qualitative comparison is in the supplementary video from minute 3:46-4:43 and supplementary Fig. A.1, where it can be observed that NeRF and REIM-NeRF exhibit poorer performance on the In-Vivo dataset, even when compared to their own results on the other datasets, likely due to the reliance on ground truth camera poses and, for REIM-NeRF, accurate depth maps. The heightened noise in In-Vivo dataset's camera poses and depth maps highlights Gaussian Pancakes' superior ability to manage data noise. However, Gaussian Pancakes exhibited a higher Depth MSE across all datasets compared to REIM-NeRF, a result of using inferred depth maps for both training and evaluation; making Depth MSE a measure of residual training error rather than a comparison with actual ground truth. Our method effectively handles data noise, prioritizing high-quality images and smooth reconstructions even with suboptimal depth references. A comparison with ColonNeRF for all datasets was not possible due to unavailable code.

\noindent\textbf{Ablation Analysis:} Table \ref{tab3} shows the incremental benefits of depth and geometric regularizations added to the foundational 3D GS method, initialized with RNNSLAM outputs, to reduce artifacts in unseen viewpoints.
However, standard metrics fall short in capturing these improvements, where test views are too close to training views to reveal the artifacts. Nevertheless, the quantitative data reveals nuanced improvements: introducing depth regularization marginally reduces Depth MSE, and normal regularization (GS+Pancaking) improves Depth SSIM, suggesting better surface consistency. The combination of both (GS+Pancaking+Depth) achieves the best overall balance, as evidenced by modest gains in PSNR and Depth SSIM, alongside a negligible increase in processing time.  Phantom and Simulation datasets show metric-based improvements that are more pronounced, further detailed in the Appendix Table A.1. For a comprehensive understanding, see the depth image comparison in supplementary Fig. A.3 and video from 4:43-5:40, which qualitatively shows artifact mitigation. Additionally, surfaces are visually smoother, seen in Fig. A.2.


\section{Conclusion}

Our evaluations across three varied datasets show that it achieves superior rendering quality, smoother reconstructions, fewer artifacts, and greatly reduces computational cost. Additionally, the method outputs explicit geometry, which sets it apart from the other leading methods with its potential for seamless integration into clinical practices. However, it's important to note that the method's success partly hinges on the quality of RNNSLAM's surface fusion, facing occasional challenges in less consistent environments. Moving forward, we plan to integrate SuperPoint \cite{barbed2022superpoint} for better point cloud creation and to refine depth map precision. This direction aims to not only bolster our method's accuracy but also its applicability in a broader range of clinical scenarios.
\\ \\
\noindent\textbf{Disclosure of Interests \& Achnowledgements}: The authors declare that they have no competing interests in this paper. This research was funded in part, by the Wellcome/EPSRC Centre for Interventional and Surgical Sciences (WEISS) [203145/Z/16/Z]; the Engineering and Physical Sciences Research Council (EPSRC) [EP/W00805X/1, EP/Y01958X/1]; Horizon 2020 FET Open [863146]; the Department of Science, Innovation and Technology (DSIT); and the Royal Academy of Engineering Chair in Emerging Technologies Scheme. 
Sierra Bonilla is supported by the UKRI AI Centre for Doctoral Training in Foundational Artificial Intelligence (FAI CDT) [EP/S021566/1].

\bibliographystyle{splncs04}
\bibliography{Paper-2298}

\appendix











\section{Appendix}

\setcounter{figure}{0}
\setcounter{table}{0}

\renewcommand{\thefigure}{A.\arabic{figure}}
\renewcommand{\thetable}{A.\arabic{table}}

\begin{figure}[ht]
\centering
\includegraphics[width=\textwidth]{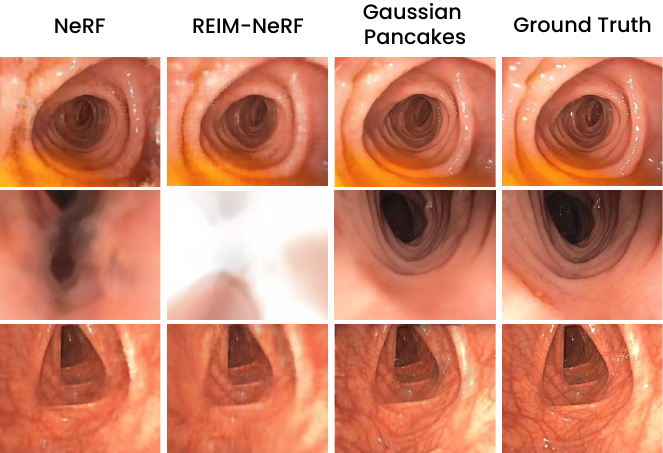}
\caption{Test images from the In-Vivo dataset showcasing the artifacts that arise in other methods.} \label{figa1}
\end{figure}

\begin{table}[ht]
\caption{Full ablation study showing effect of systematically adding all changes to the basic 3D GS method.}\label{ablationtab}
\begin{tabularx}{\textwidth}{S D S S S S}
\toprule
Dataset & Method & PSNR ↑ & \specialcell{Depth ↑\\SSIM} & \specialcell{Depth ↓\\MSE} &  \specialcell{GPU ↓\\min}\\
\toprule
\multirow{4}{*}{Simulation} & GS & 40.579 & 0.786 & 0.017 & $\bm{0.806}$ \\
& GS+Depth& $\bm{40.789}$ & 0.782 & 0.011 & 0.810 \\
& GS+Pancaking& 40.285 & 0.814 & 0.008 & 0.837 \\
& GS+Pancaking+Depth (ours) & 40.336 & $\bm{0.815}$ & $\bm{0.007}$ & 0.832 \\
\addlinespace
\multirow{4}{*}{Phantom} & GS & 32.091 & 0.811 & 1.869 & $\bm{1.078}$ \\
& GS+Depth & 32.394 & 0.813 & 1.728 & 1.368 \\
& GS+Pancaking& 32.117 & 0.868 & 0.685  & 1.124 \\
& GS+Pancaking+Depth (ours) & $\bm{32.306}$ & $\bm{0.873}$ & $\bm{0.498}$ & 1.703 \\
\addlinespace
\multirow{4}{*}{In-vivo} & GS & 26.116 &  0.444 & 0.158 & 1.211 \\
& GS+Depth & 26.062 & 0.446 & $\bm{0.145}$ & $\bm{1.206}$\\
& GS+Pancaking & 26.211 & $\bm{0.460}$ & 0.154 & 1.329 \\
& GS+Pancaking+Depth (ours) & $\bm{26.248}$ & 0.458 & 0.156 & 1.249\\
\toprule
\end{tabularx}
\end{table}

\begin{figure}[ht]
\centering
\includegraphics[width=\textwidth]{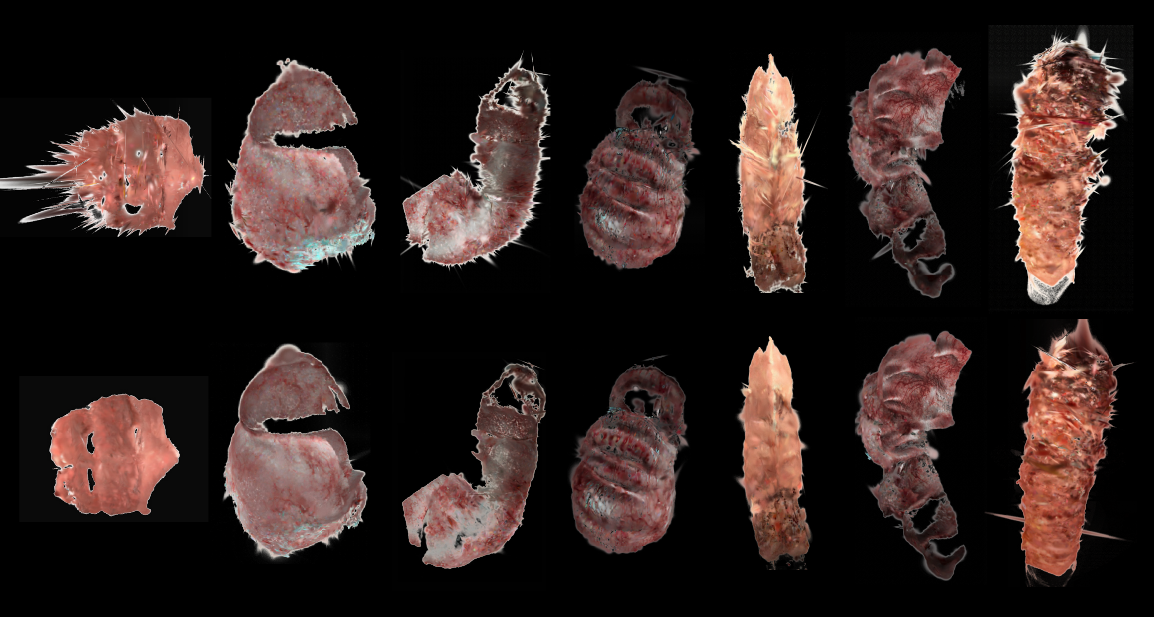}
\caption{Image showing the surface reconstruction from the basic 3D GS method (top) and from Gaussian Pancakes (bottom).} \label{figa2}
\end{figure}

\begin{figure}[ht]
\centering
\includegraphics[width=\textwidth]{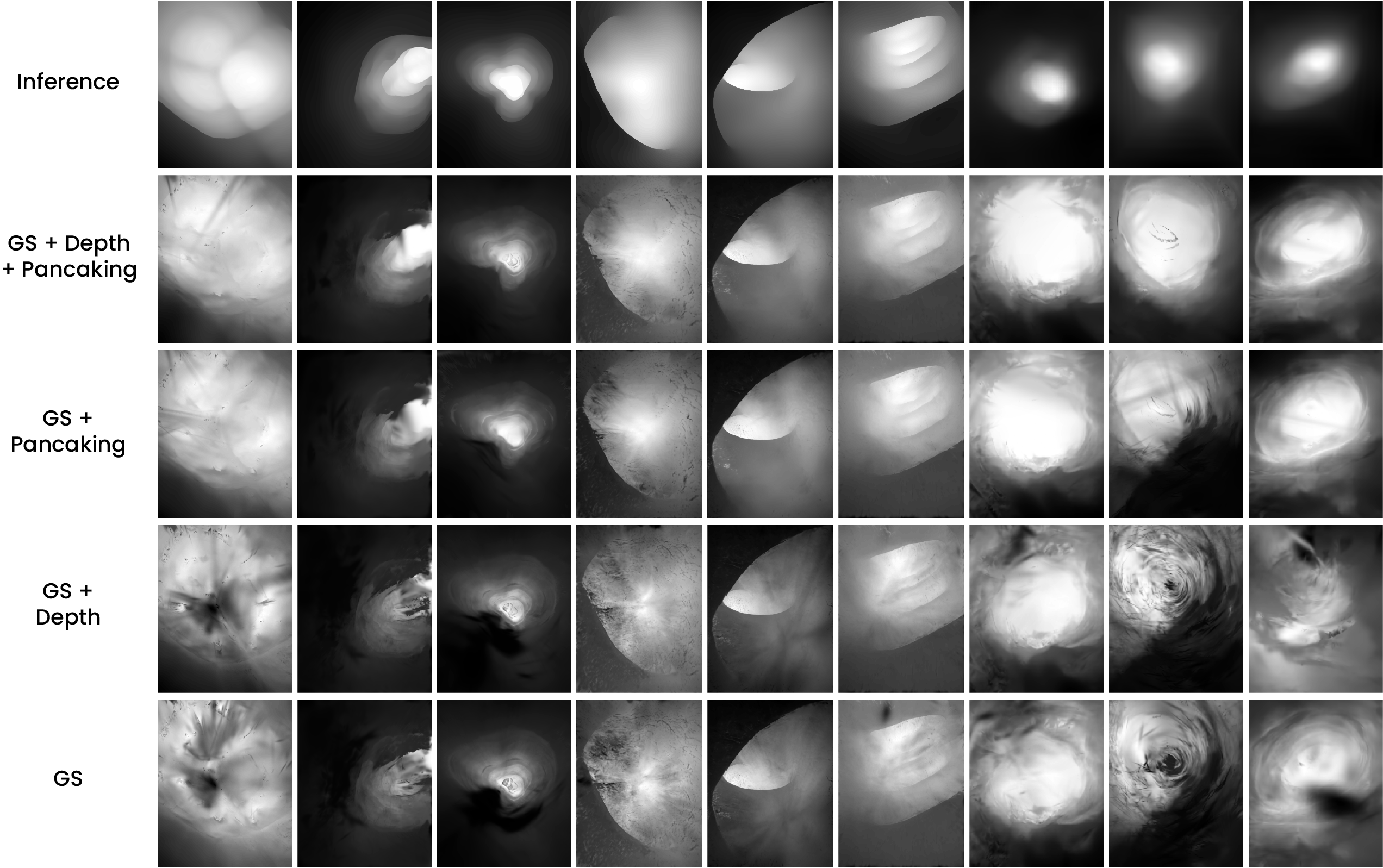}
\caption{Depth renderings showing effect of systematically adding all changes to the basic 3D GS method.} \label{figa3}
\end{figure}


\end{document}